# Điều khiển ổn định tiệm cận bền vững trong miền thời gian rời rạc cho robot di động hai bánh vi sai

## Robust asymptotic stability of two-wheels differential drive mobile robot

**Nguyễn Thị Thanh Vân, Phùng Mạnh Dương, Trần Quang Vinh**

Trường Đại học Công nghệ, ĐHQGHN

vanntt@vnu.edu.vn

**Tóm tắt**

Bài báo đề xuất bộ điều khiển ổn định bền vững cho robot di động hai bánh vi sai trong miền thời gian rời rạc dưới ảnh hưởng của nhiễu hệ thống và nhiễu phép đo. Bộ điều khiển được thực hiện dựa trên việc chia cấu hình hoạt động của robot thành hai phần: cấu hình toàn cục và cấu hình cục bộ; luật điều khiển được thiết kế theo tiêu chuẩn ổn định của Lyapunov dựa trên đặc điểm của từng phần đó. Việc áp dụng luật điều khiển đề xuất này cho phép điều khiển robot ổn định tiệm cận về đích từ bất kỳ tư thế nào ngay cả khi hệ thống chịu ảnh hưởng bởi nhiễu. Các kết quả mô phỏng đã chứng minh được hiệu quả hoạt động của phương pháp đề xuất.

**Từ khóa**: điều khiển chuyển động, ổn định Lyapunov, nhiễu hệ thống và đo, robot di động hai bánh vi sai, thời gian rời rạc

**Abstract:** The paper proposes the stable motion control law design method for non-honomic differential-drive mobile robot with system and measurement noise in discrete time domain. This method is performed basing on dividing operating configuration of robot into two parts: glocal and local configuration then the control law is designed following Lyapunov stable theory for two configuration. The proposed stable control laws is able to reach asymptotically stably to target position and orientation from any initial conditions even existing noise in the system. Some simulation results have demonstrated the effect of proposed method.

**Keywords:** motion control, Lyapunov stable theory, system and measurement noise, non-honomic differential-drive mobile robot, discrete time domain.

## 1. Phần mở đầu

Điều khiển chuyển động là khâu cuối cùng bên cạnh khối cảm nhận, định vị, lập kế hoạch của một hệ thống dẫn đường cho robot di động. Thông thường điều khiển chuyển động được chia làm hai loại: điều khiển bám theo một quỹ đạo cho trước và điều khiển điểm-điểm [1]. Điều khiển điểm-điểm hay còn được xem là điều khiển chuyển động ổn định khi yêu cầu robot phải đạt tới một tư thế (vị trí và góc) cho trước từ bất kỳ tư thế nào. Vấn đề này đã trở thành những thách thức do nhiều khó khăn trong mô hình động học của robot, đặt biệt đối với các hệ phi holonomic( hệ thống với các ràng buộc không khả tích) như mô hình của robot di động hai bánh vi sai trong hệ tọa độ Đề-các. Theo lý thuyết của Brokett [2], một hệ phi holomic không thể tồn tại luật điều khiển trạng thái tĩnh bất biến trơn và liên tục theo thời gian. Vì thế một số các nghiên cứu đã đưa ra luật điều khiển rời rạc và thay đổi theo thời gian [3-5] để áp dụng cho mô hình robot di động hai bánh vi sai biểu diễn trong hệ tọa độ Đề-các. Tuy nhiên, khi Badreddin và Mansour [6] đưa ra một phương pháp chuyển đổi việc biểu diễn mô hình từ hệ tọa độ Đề-các sang hệ tọa độ cực thông qua các biến dẫn đường thì vấn đề thiết kế các luật điều khiển ổn định phản hồi trạng thái liên tục thỏa mãn điều kiện của Brokett đã được giải quyết. Từ đó, nhiều nghiên cứu đã được đề xuất để đưa ra luật điều khiển ổn định phản hồi liên tục [7-9]. Tuy nhiên, các kết quả trên đều phân tích hệ trên miền thời gian liên tục và trong điều kiện lý tưởng không bị ảnh hưởng bởi nhiễu. Trên thực tế, vấn đề điều khiển robot được thực hiện bằng vi xử lý hoặc máy tính và nhiễu do hệ cơ khí và thiết bị đo là điều không thể tránh khỏi. Do đó việc thiết kế luật điều khiển rời rạc có tính đến ảnh hưởng của các nhiễu trong hệ thống cần phải đặc biệt quan tâm.

Trong nghiên cứu này, chúng tôi đề xuất một phương pháp thiết kế luật điều khiển chuyển động ổn định bền vững rời rạc cho mô hình rời rạc của robot di động hai bánh vi sai với ảnh hưởng của nhiễu hệ thống và nhiễu đo. Cấu hình hoạt động của robot được chia thành hai phần: cấu hình toàn cục ở xa đích và cấu hình cục bộ ở gần đích. Luật điều khiển được thiết kế cho riêng từng cấu hình cho phép điều khiển robot nhanh chóng từ cấu hình toàn cục đạt tới cấu hình cục bộ, sau đó sẽ ổn định tiệm cận về đích trong cấu hình cục bộ. Việc phân chia thành hai cấu hình và thiết kế riêng luật điều khiển như trên cho phép chống lại ảnh hưởng của nhiễu mà vẫn đạt được mục tiêu đề ra.

Cấu trúc bài báo bao gồm các phần như sau: phần 2 mô tả mô hình hệ thống rời rạc với nhiễu. Phần 3 trình bày quá trình thiết kế luật điều khiển ổn định tiệm cận về đích cho cấu hình toàn cục và cấu hình cục bộ dựa trên tiêu chuẩn ổn định Lyapunov. Phần 4 là chương trình mô phỏng hoạt động của luật điều khiển cho hai cấu hình với những phân tích và đánh giá chi tiết. Một số thảo luận và hướng phát triển tiếp theo được đề cập đến trong phần 5.

## 2. Mô hình hệ thống

Phần này trình bày mô hình động học liên tục và rời rạc của robot di động hai bánh vi sai với ràng buộc không khả tích trong hệ tọa độ Đềcac và hệ tọa độ cực của các biến dẫn đường. Tiếp theo đó mô hình động học rời rạc được xem xét cụ thể trong trường hợp hệ thống bị ảnh hưởng bởi nhiễu và đây là cơ sở để sử dụng trong phần tiếp theo.

### 2.1 Mô hình robot di động hai bánh vi sai

Robot di động được xem xét trong nghiên cứu này là loại robot có hai bánh vi sai với ràng buộc không khả tích có cấu hình và tham số được thể hiện trên hình 1.

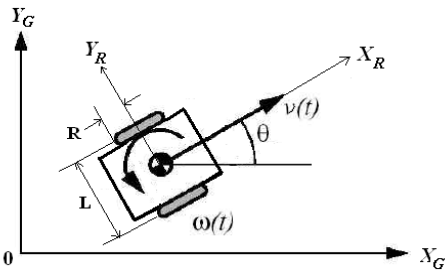

**H.1**: Mô hình robot di động hai bánh vi sai trong đó, $(X_G, Y_G)$ biểu diễn hệ tọa độ toàn cục, $(X_R, Y_R)$ biểu diễn hệ tọa độ cục bộ gắn liền với robot, $R$ ký hiệu bán kính bánh xe và $L$ là khoảng cách giữa hai bánh. Phương trình động học của robot được mô tả như sau:

$$\begin{aligned} \dot{x} &= v\cos\theta \\ \dot{y} &= v\sin\theta \\ \dot{\theta} &= \omega \end{aligned} \quad (1)$$

với $(x, y)$ là tọa độ của robot, $\theta$ là hướng của robot, $v$ và $\omega$ lần lượt là vận tốc dài và vận tốc góc của robot: $v = R(\omega_L + \omega_R)/2 \quad \omega = R(\omega_R - \omega_L)/L$ với R là bán kính bánh xe, L là khoảng cách giữa hai bánh, $\omega_L$ và $\omega_R$ lần lượt là vận tốc góc của bánh trái và bánh phải. Mô hình (1) thuộc về hệ thống cơ học có tính phi honolomo sử dụng lối vào là vận tốc do đó không thỏa mãn điều kiện của Brokett về vấn đề ổn định phản hồi. Để thỏa mãn được điều kiện của Brokett một phương pháp đã được sử dụng là chuyển đổi biểu diễn mô hình (1) trong hệ tọa độ Đềcac sang biểu diễn ở hệ tọa độ cực của các biến dẫn đường như hình 2:

$$\begin{aligned} \rho &= \sqrt{\Delta x^2 + \Delta y^2} \\ \alpha &= a\tan 2(\Delta y, \Delta x) - \theta \\ \beta &= \theta + \alpha \end{aligned} \quad (2)$$

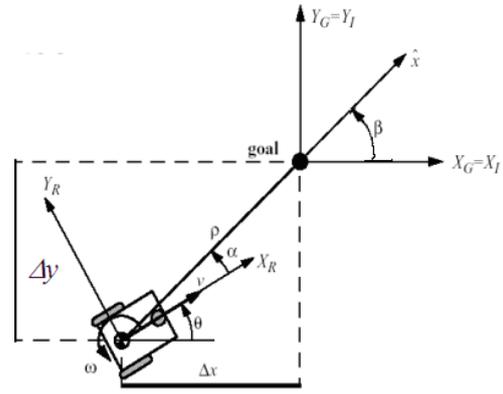

**H.2**: Robot trong không gian biến dẫn đường

Mô hình rời rạc của (1) có dạng như sau:

$$\begin{aligned} x_{i+1} &= x_i + T_s v_i \cos(\theta_i) \\ y_{i+1} &= y_i + T_s v_i \sin(\theta_i) \\ \theta_{i+1} &= \theta_i + T_s \omega_i \end{aligned} \quad (3)$$

trong đó $T_s$ là thời gian lấy mẫu của hệ thống. Phương trình động học rời rạc trong hệ tọa độ cực của biến dẫn đường theo hai khoảng giá trị của α

$$\begin{cases} \alpha = \left(-\dfrac{\pi}{2}, \dfrac{\pi}{2}\right) \\ \rho_{i+1} = \rho_i - v_i T_s \cos(\alpha_i) \\ \alpha_{i+1} = \alpha_i + v_i T_s \dfrac{\sin(\alpha_i)}{\rho_i} - \omega_i T_s \\ \beta_{i+1} = \beta_i + v_i T_s \dfrac{\sin(\alpha_i)}{\rho_i} \end{cases} \quad (4)$$

$$\begin{cases} \alpha = \left(-\pi, -\dfrac{\pi}{2}\right] \cup \left(\dfrac{\pi}{2}, \pi\right] \\ \rho_{i+1} = \rho_i + v_i T_s \cos(\alpha_i) \\ \alpha_{i+1} = \alpha_i - v_i T_s \dfrac{\sin(\alpha_i)}{\rho_i} - \omega_i T_s \\ \beta_{i+1} = \beta_i - v_i T_s \dfrac{\sin(\alpha_i)}{\rho_i} \end{cases} \quad (5)$$

### 2.2 Mô hình rời rạc có nhiễu

Sơ đồ khối hệ thống điều khiển chuyển động với ảnh hưởng của nhiễu lối vào và nhiễu phản hồi được trình bày ở hình 3.

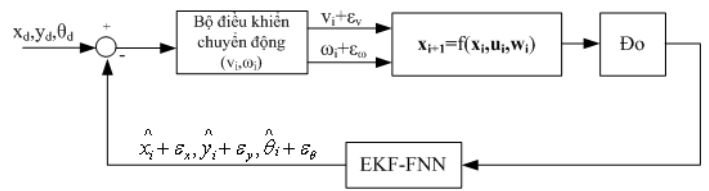

**H.3**: Sơ đồ hệ thống điều khiển chuyển động có nhiễu

Trong sơ đồ hệ thống trên thì EKF-FNN là bộ lọc đã được nhóm tác giả đề xuất trong bài toán định vị robot ở một nghiên cứu khác [11]. Giả sử tại thời điểm $i$, sai số giữa giá trị ước lượng $(\hat{x}_i, \hat{y}_i, \hat{\theta}_i)$ từ bộ lọc EKF-FNN với giá trị thực chính là nhiễu ước lượng hay còn gọi là nhiễu phép đo. Gọi nhiễu đ này là các đại lượng $|\varepsilon_X| \leq \varepsilon_X^{max}, |\varepsilon_Y| \leq \varepsilon_Y^{max}, |\varepsilon_\theta| \leq \varepsilon_\theta^{max}$. Để mang tính tổng quát, gọi giá trị đo này là $(X_i, Y_i, \theta_i)$, khi đó giá trị phản hồi có nhiễu sẽ là $\hat{X}_i = X_i + \varepsilon_X, \hat{Y}_i = Y_i + \varepsilon_Y, \hat{\theta}_i = \theta_i + \varepsilon_\theta$. Các biến dẫn đường với sẽ chịu tác động của nhiễu sau:

$$\varepsilon_\rho = \sqrt{(X_d - \hat{X}_i)^2 + (Y_d - \hat{Y}_i)^2} - \sqrt{(X_d - X_i)^2 + (Y_d - Y_i)^2}$$

$$\varepsilon_\alpha = a\tan2(Y_d - \hat{Y}_i, X_d - \hat{X}_i) - a\tan2(Y_d - Y_i, X_d - X_i)$$

$$\varepsilon_\beta = \varepsilon_\alpha + \varepsilon_\theta$$

Vận tốc lối vào bị ảnh hưởng bởi nhiễu hệ thống ký hiệu là $|\varepsilon_v| \leq \varepsilon_v^{max}, |\varepsilon_\omega| \leq \varepsilon_\omega^{max}$ do đó lối vào có nhiễu sẽ là $\hat{v}_i = v_i + \varepsilon_v, \hat{\omega}_i = \omega_i + \varepsilon_\omega$. Phương trình động học điều khiển chuyển động trong trường hợp có nhiễu hệ thống và nhiễu đo của biến dẫn đường (4) và (5) trở thành:

$$\begin{cases} \alpha = \left(-\dfrac{\pi}{2}, \dfrac{\pi}{2}\right] \\ \rho_{i+1} = \rho_i - (v_i + \varepsilon_v)T_s \cos(\alpha_i) \\ \alpha_{i+1} = \alpha_i + (v_i + \varepsilon_v)T_s \dfrac{\sin(\alpha_i)}{\rho_i} - (\omega_i + \varepsilon_\omega)T_s \\ \beta_{i+1} = \beta_i + (v_i + \varepsilon_v)T_s \dfrac{\sin(\alpha_i)}{\rho_i} \end{cases} \quad (6)$$

$$\begin{cases} \alpha = \left(-\pi, -\dfrac{\pi}{2}\right] \cup \left(\dfrac{\pi}{2}, \pi\right] \\ \rho_{i+1} = \rho_i + (v_i + \varepsilon_v)T_s \cos(\alpha_i) \\ \alpha_{i+1} = \alpha_i - (v_i + \varepsilon_v)T_s \dfrac{\sin(\alpha_i)}{\rho_i} - (\omega_i + \varepsilon_\omega)T_s \\ \beta_{i+1} = \beta_i - (v_i + \varepsilon_v)T_s \dfrac{\sin(\alpha_i)}{\rho_i} \end{cases} \quad (7)$$

với $v_i = f(v_i, \varepsilon_\rho, \varepsilon_\alpha, \varepsilon_\beta)$ và $\omega_i = f(\omega_i, \varepsilon_\rho, \varepsilon_\alpha, \varepsilon_\beta)$.

## 3. Thiết kế bộ điều khiển ổn định rời rạc

Mục đích của bộ điều khiển chuyển động là thiết kế luật điều khiển ($v_i$, $\omega_i$) sao cho điều khiển robot ổn định tiệm cận về đích dựa trên tư thế robot và nhiễu tại thời điểm hiện tại $i$. Phần này sẽ trình bày bộ điều khiển với mục tiêu trên như sau: thứ nhất cấu hình của robot trong hệ tọa độ cực của các biến dẫn đường được chia thành hai cấu hình: cấu hình toàn cục ($\Omega_G$) là ở xa cấu hình đích, cấu hình cục bộ ($\Omega_L$) là ở gần cấu hình đích; thứ hai luật điều khiển trong từng cấu hình được thiết kế dựa trên lý thuyết ổn định của Lyapunov. Trong cấu hình toàn cục, luật điều khiển được thiết kế sao cho robot đạt tới cấu hình cục bộ một cách nhanh nhất có thể trong khi ở cấu hình cục bộ luật điều khiển phải có khả năng loại bỏ được nhiễu và tiệm cận ổn định về đích.

### 3.1 Thiết kế luật điều khiển trong cấu hình $\Omega_G$

**A. Trường hợp phương trình động học (6)**

Chọn hàm Lyapunov rời rạc tại thời điểm $i$

$$V_i(\rho_i, \alpha_i, \beta_i) = \frac{1}{2}\rho_i^2 + \frac{1}{2}\alpha_i^2 + \frac{1}{2}h\beta_i^2 > 0, \forall h > 0 \quad (8)$$

Sử dụng khái niệm đạo hàm rời rạc $V_i$ theo sai phân tiến cho (8) ta thu được đạo hàm Lyapunov rời rạc như sau:

$$\Delta V_i = V_{i+1} - V_i = \frac{1}{2}2\rho_i\Delta\rho_i + \frac{1}{2}2\alpha_i\Delta\alpha_i + \frac{1}{2}2h\beta_i\Delta\beta_i$$
$$= \rho_i(\rho_{i+1} - \rho_i) + \alpha_i(\alpha_{i+1} - \alpha_i) + h\beta_i(\beta_{i+1} - \beta_i) \quad (9)$$

Chọn luật điều khiển vận tốc dài và vận tốc góc bị ảnh hưởng bởi nhiễu phản hồi có dạng như sau:

$$v_i = \gamma \tanh(\rho_i + \varepsilon_\rho)\cos(\alpha_i + \varepsilon_\alpha)$$

$$\omega_i = k(\alpha_i + \varepsilon_\alpha) + \gamma \frac{\sin(\alpha_i + \varepsilon_\alpha)}{(\alpha_i + \varepsilon_\alpha)} \frac{\tanh(\rho_i + \varepsilon_\rho)}{(\rho_i + \varepsilon_\rho)} \quad (10)$$

$$\cos(\alpha_i + \varepsilon_\alpha)\left[(\alpha_i + \varepsilon_\alpha) + h(\beta_i + \varepsilon_\beta)\right]$$

Việc chọn luật điều khiển $v_i$ có dạng như trên cho phép hạn chế tốc độ cực đại của robot bằng tham số $\gamma = v_{max}$. Thay (6) và (10) vào (9) thu được các kết quả sau:

$$\Delta V_i(1) = \rho_i(\rho_{i+1} - \rho_i) =$$
$$-\gamma\rho_i T_s \tanh(\rho_i + \varepsilon_\rho)\cos(\alpha_i)\cos(\alpha_i + \varepsilon_\alpha) - \varepsilon_v \rho_i T_s \cos(\alpha_i)$$

vì

$$\alpha, \alpha + \varepsilon_\alpha \in \left(-\pi/2, \pi/2\right) \rightarrow \cos(\alpha_i) > 0, \cos(\alpha_i + \varepsilon_\alpha) > 0$$

$$\rho > 0, \varepsilon_\rho > 0 \rightarrow \tanh(\rho + \varepsilon_\rho) > 0$$

nên chọn $\gamma$ đủ lớn để loại bỏ thành phần thứ 2 chứa nhiễu $\varepsilon_v$ để $\Delta V_i(1) \leq 0$, do đó $\rho$ sẽ hội tụ đến giá trị nhỏ.

$$\Delta V_i(2) = \alpha_i(\alpha_{i+1} - \alpha_i) + h\beta_i(\beta_{i+1} - \beta_i)$$

$$= -k\alpha_i^2 T_s$$

$$+ \gamma \alpha_i T_s \frac{\tanh(\rho_i + \varepsilon_\rho)}{\rho_i} \sin(\alpha_i)\cos(\alpha_i + \varepsilon_\alpha)$$

$$- \gamma \alpha_i T_s \frac{\tanh(\rho_i + \varepsilon_\rho)}{(\rho_i + \varepsilon_\rho)} \sin(\alpha_i + \varepsilon_\alpha)\cos(\alpha_i + \varepsilon_\alpha)$$

$$+ \gamma h \beta_i T_s \frac{\tanh(\rho_i + \varepsilon_\rho)}{\rho_i} \sin(\alpha_i)\cos(\alpha_i + \varepsilon_\alpha)$$

$$- \gamma h \beta_i T_s \frac{\alpha_i}{(\alpha_i + \varepsilon_\alpha)} \frac{\tanh(\rho_i + \varepsilon_\rho)}{(\rho_i + \varepsilon_\rho)} \sin(\alpha_i + \varepsilon_\alpha)\cos(\alpha_i + \varepsilon_\alpha)$$

$$- \varepsilon_\beta h T_s \gamma \frac{\alpha_i}{(\alpha_i + \varepsilon_\alpha)} \frac{\tanh(\rho_i + \varepsilon_\rho)}{(\rho_i + \varepsilon_\rho)} \sin(\alpha_i + \varepsilon_\alpha)\cos(\alpha_i + \varepsilon_\alpha)$$

$$+ \varepsilon_v \alpha_i T_s \frac{\sin(\alpha_i)}{\rho_i} - \varepsilon_\omega \alpha_i T_s + \varepsilon_v h \beta_i T_s \frac{\sin(\alpha_i)}{\rho_i} - \varepsilon_\alpha k \alpha_i T_s$$

Vì

$$\alpha, \alpha + \varepsilon_\alpha \in (-\pi/2, \pi/2)$$

$$\Leftrightarrow \begin{cases} |\alpha \sin(\alpha + \varepsilon_\alpha)| > |\alpha \sin \alpha| & khi \quad \varepsilon_\alpha > 0 \\ |\alpha \sin(\alpha + \varepsilon_\alpha)| < |\alpha \sin \alpha| & khi \quad \varepsilon_\alpha < 0 \end{cases}$$

$$\frac{\tanh(\rho + \varepsilon_\rho)}{\rho} > \frac{\tanh(\rho + \varepsilon_\rho)}{(\rho + \varepsilon_\rho)} > 0$$

nên

$$\gamma \alpha_i T_s \frac{\tanh(\rho_i + \varepsilon_\rho)}{\rho_i} \sin(\alpha_i)\cos(\alpha_i + \varepsilon_\alpha)$$

$$- \gamma \alpha_i T_s \frac{\tanh(\rho_i + \varepsilon_\rho)}{(\rho_i + \varepsilon_\rho)} \sin(\alpha_i + \varepsilon_\alpha)\cos(\alpha_i + \varepsilon_\alpha)$$

và

$$\gamma h \beta_i T_s \frac{\tanh(\rho_i + \varepsilon_\rho)}{\rho_i} \sin(\alpha_i)\cos(\alpha_i + \varepsilon_\alpha)$$

$$- \gamma h \beta_i T_s \frac{\alpha_i}{(\alpha_i + \varepsilon_\alpha)} \frac{\tanh(\rho_i + \varepsilon_\rho)}{(\rho_i + \varepsilon_\rho)} \sin(\alpha_i + \varepsilon_\alpha)\cos(\alpha_i + \varepsilon_\alpha)$$

VCM-2014

cũng tiến tới giá trị nhỏ ε hoặc –ε nào đó. Chọn k đủ lớn để loại bỏ được các giá trị nhỏ của hai thành phần trên và các thành phần liên quan đến nhiễu nhỏ để $\Delta V_i(2) \leq 0$, do đó α và β sẽ hội tụ đến giá trị nhỏ. Thay (10) vào (6) ta được phương trình động học của hệ thống có dạng:

$$\begin{cases} \rho_{i+1} = \rho_i - \left\{ \gamma \tanh(\rho_i + \varepsilon_\rho)\cos(\alpha_i)\cos(\alpha_i + \varepsilon_\alpha) + \varepsilon_v \cos\alpha_i \right\} \\ \alpha_{i+1} = \alpha_i + \left\{ \gamma \tanh(\rho_i + \varepsilon_\rho)\cos(\alpha_i + \varepsilon_\alpha) + \varepsilon_v \right\} \frac{\sin\alpha_i}{\rho_i} \\ - \left\{ k(\alpha_i + \varepsilon_\alpha) + \gamma \frac{\sin(\alpha_i + \varepsilon_\alpha)}{(\alpha_i + \varepsilon_\alpha)} \frac{\tanh(\rho_i + \varepsilon_\rho)}{(\rho_i + \varepsilon_\rho)} \cos(\alpha_i + \varepsilon_\alpha) \right\} \\ \left[ (\alpha_i + \varepsilon_\alpha) + h(\beta_i + \varepsilon_\beta) \right] + \varepsilon_\omega \\ \beta_{i+1} = \beta_i + \left\{ \gamma \tanh(\rho_i + \varepsilon_\rho)\cos(\alpha_i + \varepsilon_\alpha) + \varepsilon_v \right\} \frac{\sin(\alpha_i)}{\rho_i} \end{cases}$$

(11)

**B. Trường hợp phương trình động học (7)**

Cũng chọn hàm $V_i$ Lyapunov rời rạc như (8) và thu được phương trình đạo hàm $\Delta V_i$ rời rạc như (9). Trong trường hợp đặt v = -v để định nghĩa lại hướng tiến của robot nên luật điều khiển (10) sẽ trở thành:

$$v_i = -\gamma \tanh(\rho_i + \varepsilon_\rho)\cos(\alpha_i + \varepsilon_\alpha)$$

$$\omega_i = k(\alpha_i + \varepsilon_\alpha) + \gamma \frac{\sin(\alpha_i + \varepsilon_\alpha)}{(\alpha_i + \varepsilon_\alpha)} \frac{\tanh(\rho_i + \varepsilon_\rho)}{(\rho_i + \varepsilon_\rho)} \quad (12)$$

$$\cos(\alpha_i + \varepsilon_\alpha)\left[(\alpha_i + \varepsilon_\alpha) + h(\beta_i + \varepsilon_\beta)\right]$$

Thay (12) và (7) vào (9) ta thu được

$$\Delta V_i(1) = \rho_i(\rho_{i+1} - \rho_i) =$$

$$-\gamma \rho_i T_s \tanh(\rho_i + \varepsilon_\rho)\cos(\alpha_i)\cos(\alpha_i + \varepsilon_\alpha) + \varepsilon_v \rho_i T_s \cos(\alpha_i)$$

vì

$$\alpha, \alpha + \varepsilon_\alpha \in (-\pi, -\pi/2] \cup (\pi/2, \pi)$$

$$\rightarrow \cos(\alpha_i) < 0, \cos(\alpha_i + \varepsilon_\alpha) < 0$$

$$\rho_i > 0, \varepsilon_\rho > 0 \rightarrow \tanh(\rho_i + \varepsilon_\rho) > 0$$

nên với γ đủ lớn như ở (10) đủ để loại bỏ thành phần chứa thứ 2 chứa nhiễu để $\Delta V_i(1) \leq 0$, do đó ρ cũng hội tụ đến giá trị nhỏ. Tương tự ta thu được $\Delta V_i(2)$ gần giống với ở trường hợp A chỉ khác dấu ở thành phần $\alpha_i\varepsilon_v T_s\sin(\alpha_i)/\rho_i$ và $\varepsilon_v h\beta_i T_s\sin(\alpha_i)/\rho_i$. Tuy nhiên đây là những thành phần liên quan đến nhiễu có giá trị nhỏ nên biện luận giống trường hợp A ta cũng có được $\Delta V_i(2) \leq 0$ do đó α và β cũng sẽ hội tụ đến giá trị nhỏ. Thay (12) vào (7) ta cũng thu được phương trình động học của hệ thống giống (11). Như vậy luật điều khiển chống lại nhiễu hệ thống và nhiễu đo trong cấu hình toàn cục được viết lại như sau:

$$\begin{cases} v_i = \gamma \tanh \rho_i \cos \alpha_i \\ \omega_i = k\alpha_i + \gamma \dfrac{\sin \alpha_i}{\alpha_i} \dfrac{\tanh \rho_i}{\rho_i} \cos \alpha_i \left[ \alpha_i + h\beta_i \right] \\ \alpha = \left( -\dfrac{\pi}{2}, \dfrac{\pi}{2} \right] \end{cases} \quad (13)$$

$$\begin{cases} v_i = -\gamma \tanh \rho_i \cos \alpha_i \\ \omega_i = k\alpha_i + \gamma \dfrac{\sin \alpha_i}{\alpha_i} \dfrac{\tanh \rho_i}{\rho_i} \cos \alpha_i \left[ \alpha_i + h\beta_i \right] \\ \alpha = \left( -\pi, -\dfrac{\pi}{2} \right] \cup \left( \dfrac{\pi}{2}, \pi \right] \end{cases}$$

### 3.2 Thiết kế luật điều khiển trong cấu hình $\Omega_L$

Như đã phân tích trong 3.1 luật điều khiển (13) cho phép robot đạt đến các giá trị nhỏ, các giá trị nhỏ này chính là cấu hình $\Omega_L$. Giả sử các giá trị nhỏ đạt được là $\rho = \varepsilon_P, \alpha \to \varepsilon_\alpha, \beta \to \varepsilon_\beta$, khi đó (4.11) trở thành:

$$\begin{cases} \rho_{i+1} = \rho_i - \left\{ \gamma(\varepsilon_P + \varepsilon_\rho) + \varepsilon_v \right\} \\ \alpha_{i+1} = \alpha_i + \gamma\varepsilon_\alpha(1 + \dfrac{\varepsilon_\rho}{\varepsilon_P}) - \left\{ 2\varepsilon_\alpha(k+\gamma) + 2\gamma h\varepsilon_\beta + \varepsilon_\omega \right\} \\ \beta_{i+1} = \beta_i + \gamma\varepsilon_\alpha(1 + \dfrac{\varepsilon_\rho}{\varepsilon_P}) \end{cases} \quad (14)$$

Nếu chọn hàm Lyapunov và luật điều khiển cho cấu hình $\Omega_L$ giống như (13) của cấu hình $\Omega_G$ thì

$$\Delta V_i(1) = \rho_i(\rho_{i+1} - \rho_i) = \left\{ \gamma(-\varepsilon_P^2 - \varepsilon_P \varepsilon_\rho) - \varepsilon_P \varepsilon_v \right\}$$

Nếu chọn $\varepsilon_P > \left| \varepsilon_\rho \right| + \left| \varepsilon_v \right| / \gamma$ thì $\Delta V_i(1) < 0$, do đó $\rho$ hội tụ về 0, vì thế luật điều khiển v của (13) của cấu hình $\Omega_G$ vẫn áp dụng được cho cấu hình $\Omega_L$. Tuy nhiên với

$$\Delta V_i(2) = \beta_i(\beta_{i+1} - \beta_i) = \varepsilon_\beta \gamma \varepsilon_\alpha (1 + \dfrac{\varepsilon_\rho}{\varepsilon_P}) > 0 \quad khi \quad \varepsilon_\alpha > 0, \varepsilon_\beta > 0$$

do đó $\beta$ sẽ không hội tụ hay hệ trở nên mất ổn định trong một vùng giá trị nào đó của nhiễu đo. Vì giá trị của $\alpha$ trong (14) tỉ lệ với $\beta$ nên cũng có thể xảy ra trường hợp tương tự. Cần xây dựng lại luật điều khiển cho $\omega$ sao cho giá trị $\alpha$ và $\beta$ hội tụ về 0. Theo (2), $\theta = \beta - \alpha$ nên chọn lại hàm Lyapunov và luật điều khiển cho $\omega$ như sau:

$$V_i = \dfrac{1}{2}\rho_i^2 + \dfrac{1}{2}(\beta_i - \alpha_i)^2 = \dfrac{1}{2}\rho_i^2 + \dfrac{1}{2}\theta_i^2 > 0 \quad (15)$$
$$\omega_i = -k_2 \theta_i$$

Khi đó

$$\Delta V_i = \rho_i(\rho_{i+1} - \rho_i) + \theta_i(\theta_{i+1} - \theta_i) = \left\{ \gamma(-\varepsilon_P^2 - \varepsilon_P\varepsilon_\rho) - \varepsilon_P\varepsilon_v \right\} - k_2\theta_i^2 - \varepsilon_\omega k_2 \theta_i \quad (16)$$

Việc sử dụng biến $\theta = \beta - \alpha$ cho phép loại bỏ được nhiễu đo và nhiễu hệ thống của $\omega_i$ ở (16). Chọn $k_2$ đủ lớn để loại bỏ $\varepsilon_\omega$ sao cho $\Delta V_i < 0$, khi đó $\rho$, $\theta$ hay $\beta-\alpha$ sẽ hội tụ tiệm cận về 0.

**Tóm lại**: Luật điều khiển chuyển động ổn định tiệm cận về đích của robot trong trường hợp chịu ảnh hưởng của nhiễu hệ thống và nhiễu phép đo nhỏ trong hai cấu hình như sau:

$$v_i = \begin{cases} \gamma \tanh(\rho_i) \cos(\alpha_i) & \alpha = \alpha_1 \\ -\gamma \tanh(\rho_i) \cos(\alpha_i) & \alpha = \alpha_2 \end{cases}$$

$$\omega_i = \begin{cases} k\alpha_i + \gamma \dfrac{\sin(\alpha_i)}{\alpha_i} \dfrac{\tanh(\rho_i)}{\rho_i} \cos(\alpha_i)\left[\alpha_i + h\beta_i\right], \Omega_G \\ -k_2\theta_i & \Omega_L \end{cases} \quad (17)$$

với $\alpha_1 \in$ (6), $\alpha_2 \in$ (7).

## 4. Mô phỏng

### 4.1 Cài đặt mô phỏng

Chương trình mô phỏng được thực hiện để đánh giá hiệu quả của luật điều khiển với hai cấu hình trong trường hợp hệ thống chịu ảnh hưởng của nhiễu. Các tham số mô phỏng được thiết lập dựa trên robot thực tế tại phòng thí nghiệm [12], cụ thể như sau:

- Đường kính của bánh xe: R = 0,05 *m*.
- Khoảng cách giữa hai bánh xe: L = 0,6 *m*.
- Vận tốc cực đại của robot: $v_{max}$ = 1,3 *m/s*.
- Thời gian lấy mẫu của hệ thống: $T_S$ = 100 *ms*.
- Sai số cho phép của hệ thống khi về đích là khoảng cách $\rho = 10^{-5}$ *m*.
- Giá trị các tham số cho bộ điều khiển được chọn như sau: $\gamma$ = 1,3; $k$ = 1; $h$ = 0,17; $k_2$ = 2,7.
- Các giá trị nhiễu đo được lựa chọn dựa trên cơ sở là sai lệch lớn nhất ở phần ước lượng của từ bộ EKF-FNN đối với robot thực: $\varepsilon_X^{max}$ = 0.3 *m*, $\varepsilon_Y^{max}$ = 0.3 *m*, $\varepsilon_\theta^{max}$ = 0.17 *rad*.
- Nhiễu hệ thống dựa trên khảo sát thực nghiệm với robot thực điều khiển motor bằng thuật toán PID với sai số của vận tốc góc $\omega_L$ và $\omega_R$ là ±5%. Vì thế với $v_{max} = 1,3$ *m/s* thì $\varepsilon_v^{max} = 0,065$ và $\varepsilon_\omega^{max} = 0,2167$. Giá trị $\varepsilon_P$ để chuyển sang cấu hình cục bộ được chọn sao cho thỏa mãn điều kiện $\varepsilon_P > |\varepsilon_\rho| + |\varepsilon_v|/\gamma$.

- Nhiễu hệ thống và nhiễu đo là các giá trị nhỏ có phân bố ngẫu nhiên trong khoảng giới hạn
$|\varepsilon_X| \leq \varepsilon_X^{max}, |\varepsilon_Y| \leq \varepsilon_Y^{max}, |\varepsilon_\theta| \leq \varepsilon_\theta^{max}$
$|\varepsilon_v| \leq \varepsilon_v^{max}, |\varepsilon_\omega| \leq \varepsilon_\omega^{max}$.

Phương pháp mô phỏng Monte-Carlo được thực hiện 100 lần cho mỗi chương trình mô phỏng để đánh giá chính xác hiệu quả của thhhhhuật toán.

**4.2 Mô phỏng**

**A. So sánh luật điều khiển trong hai cấu hình**

Chương trình mô phỏng thực hiện điều khiển chuyển động robot từ một cấu hình bất kỳ ổn định tiệm cận về đích. Để chứng minh rõ hơn hiệu quả của việc phân chia luật điều khiển trong hai cấu hình, chương trình mô phỏng sẽ thực hiện việc so sánh giữa trường hợp áp dụng luật điều khiển (13) và (17) cho cả hai cấu hình toàn cục và cục bộ. Cả hai trường hợp đều thực hiện với điều kiện robot xuất phát tại $(-2,-5.5,30^0)$ ổn định về đích $(0,0,0^0)$. Các kết quả thu được khi sử dụng luật điều khiển (13) cho cả hai cấu hình được thể hiện ở hình 4-6: mặc dù đường đi ở hình 4 ổn định về đích $(x,y) = (0.0071m, -0.0097m)$ nhưng vẫn tồn tại góc lệch $\theta = 0.7519$ rad do áp dụng luật điều khiển của cấu hình toàn cục trong cấu hình cục bộ; đáp ứng theo x, y và góc hướng θ thể hiện ở hình 5 cho thấy rõ sự biến thiên của các giá trị này khi về đích, hình 6 là đáp ứng của v và ω do ảnh hưởng bởi nhiễu ngẫu nhiên nên cả v và ω đều dao động khi ở cấu hình toàn cục nhưng khi đến cấu hình cục bộ thì chỉ còn ω dao động, hệ trở nên mất ổn định và góc θ không thể tiệm cận về 0 được. Hình 7-9 là kết quả khi áp dụng luật điều khiển (17) cho hai cấu hình: đường đi ở hình 7 đã ổn định tiệm cận về đích cho cả $(x,y,\theta) = (0.0028m, 0.1049m, 0.0066$ rad$)$ và đáp ứng của chúng ở hình 8 đã minh chứng cho điều này; đáp ứng ở hình 9 cho thấy v và ω chỉ bị dao động do nhiễu ở cấu hình toàn cục và gần như bị loại bỏ hẳn khi ở cấu hình cục bộ. Giá trị vận tốc dài v không vượt quá 1.3m/s bằng với giá trị của tham số γ. Kết quả so sánh trong hai trường

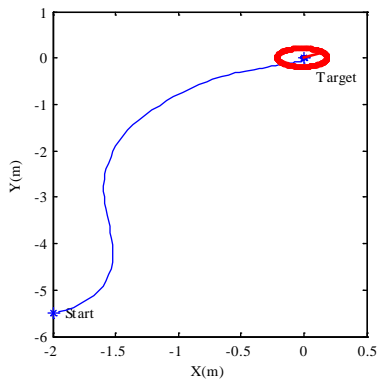 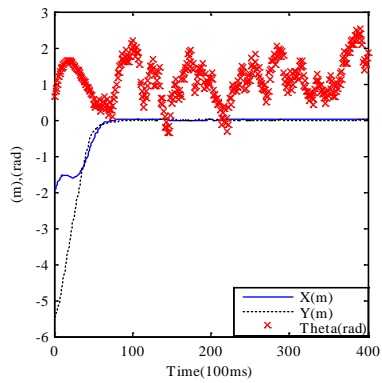 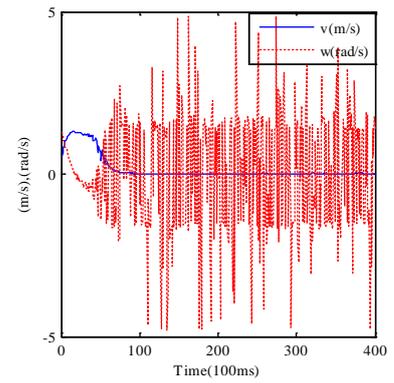

**H.4**: Đường đi theo luật điều khiển (13)   **H.5**: Đáp ứng x, y và θ theo luật (13)   **H.6**: Đáp ứng v và ω theo luật (13)

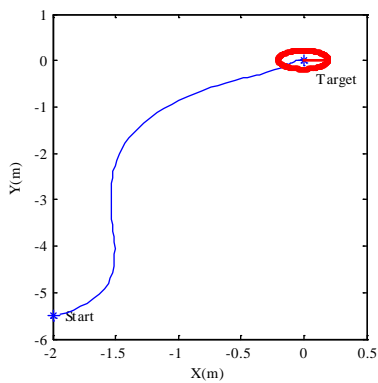 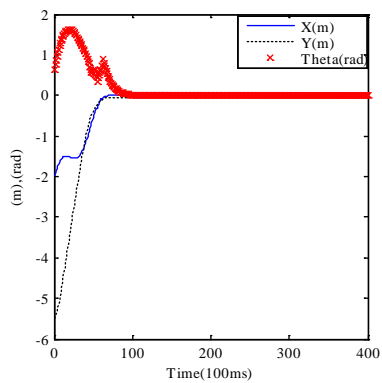 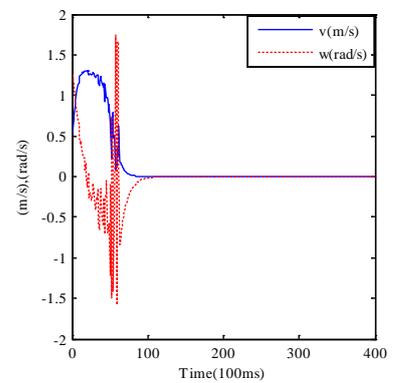

**H.7**: Đường đi theo luật điều khiển (17)   **H.8**: Đáp ứng x, y và θ theo luật (17)   **H.9**: Đáp ứng v và ω theo luật (17)

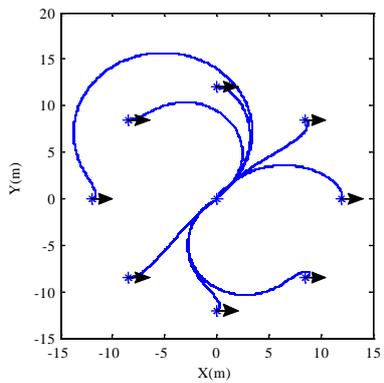 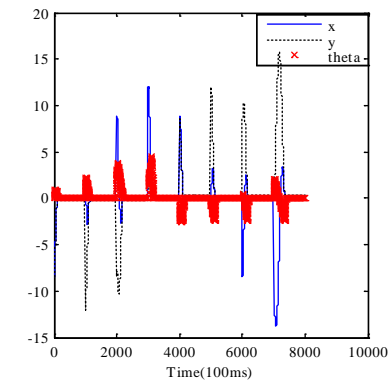 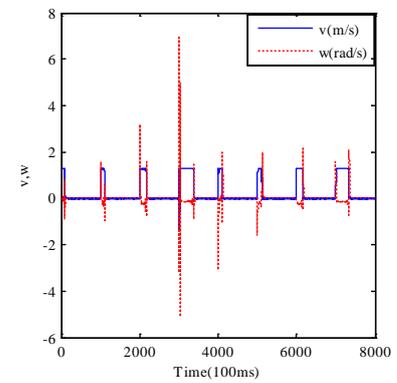

**H.10:** Đường đi tại điểm xuất phát khác nhau với cùng góc ban đầu $0^0$   **H.11**: Đáp ứng x, y và θ cho các đường đi ở H.10   **H.12**: Đáp ứng v và ω cho các đường đi ở H.10

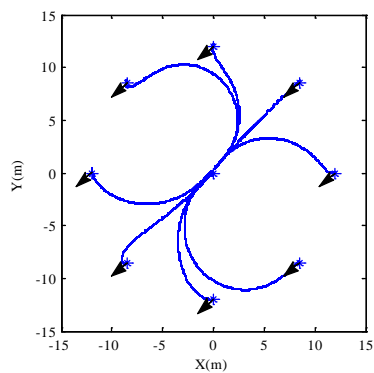 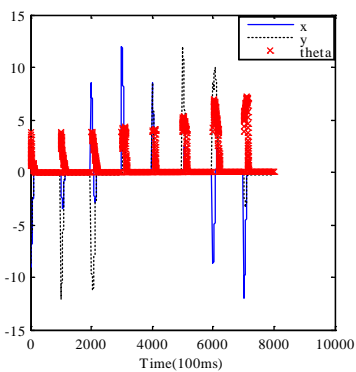 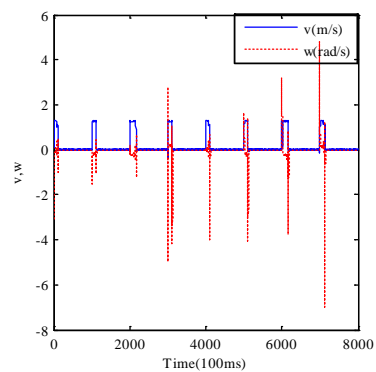

**H.13**: Đường đi tại điểm xuất phát khác nhau với cùng góc ban đầu $220^0$

**H.14**: Đáp ứng x, y và θ cho các đường đi ở H.13

**H.15**: Đáp ứng v và ω cho các đường đi ở H.13

hợp trên đã khẳng định rằng nếu chỉ sử dụng một luật điều khiển trong cấu hình toàn cục cho cả cấu hình cục bộ khi rất gần đích thì hệ thống không thể tiệm cận về đích do ảnh hưởng bởi nhiễu. Trong khi sử dụng luật điều khiển cho hai cấu hình toàn cục và cục bộ riêng biệt như phương pháp đề xuất đã chống lại được nhiễu và ổn định tiệm cận về đích.

**B. Mô phỏng với nhiều tư thế xuất phát khác nhau**

Để đánh giá luật điều khiển (17) cho cả hai khoảng giá trị của α sao cho robot luôn chuyển động theo hướng tiến, chương trình thực hiện mô phỏng với các điểm xuất phát khác nhau nằm trên đường tròn bán kính R=12m, khi đó góc α sẽ nằm trong cả hai khoảng $α_1$ và $α_2$. Hình 10 thể hiện đường đi tại 8 điểm cách nhau một góc π/4 với cùng góc xuất phát $0^0$. Đáp ứng theo x,y và θ ở hình 11 và đáp ứng v, ω ở hình 12. Hình 13 là 8 đường đi khác với góc xuất phát $220^0$ và các đáp ứng tương ứng ở hình 14 và 15. Từ các kết quả trên ta có thể thấy rằng luật điều khiển (17) đã điều khiển robot chuyển động ổn định tiệm cận về đích từ bất kỳ tư thế nào trong trường hợp hệ thống chịu ảnh hưởng của nhiễu hệ thống và nhiễu đo. Các kết quả này đã chứng minh tính hiệu quả của phương pháp đề xuất trong phạm vi hoạt động rộng của robot khi đang ở bất kỳ tư thế nào đều có thể ổn định tiệm cận về đích. Khả năng này cho phép robot có thể hoạt động tốt trong môi trường thực.

## 5. Kết luận

Đóng góp chính của báo cáo là đã xây dựng thành cộng luật điều khiển chuyển động ổn định tiệm cận bền vững cho robot di động hai bánh vi sai trong trường hợp hệ thống bị ảnh hưởng bởi nhiễu. Luật điều khiển được thiết kế theo lý thuyết ổn định của Lyapunov cho hai cấu hình hoạt động: cấu hình toàn cục khi ở xa đích và cấu hình cục bộ khi ở rất gần đích. Việc phân chia thành hai cấu hình cùng với luật điều khiển tương ứng cho phép điều khiển robot ổn định tiệm cận về đích ngay cả khi có nhiễu. Hiệu quả hoạt động của phương pháp đề xuất đã được chứng minh qua chương trình mô phỏng với những khảo sát và đánh giá kết quả một cách chi tiết. Trong nghiên cứu tiếp theo, phương pháp này sẽ ứng dụng trong hệ thống dẫn đường tự động của robot di động với nhiệm vụ là hành vi về đích.

Hội nghị toàn quốc lần thứ 7 về Cơ điện tử - VCM-2014

vehicles via Lyapunov techniques, IEEE Robot. & Autom. Mag., 2 (1) (1995) 27-35

[8] M. Kim and P. Tsiotras, - Controllers for unicycle-type wheeled robots: Theoretical results and experimental validation, IEEE Trans. Robot. & Autom., 18 (3) (2002) 294-307

[9] T. Jin and H.H.Tack- Path following control of mobile robot using Lyapunov techniques and PID controller, International Journal ò Fuzzy Logic and Intelligent Systems, vol 11, no.1, March 2011, pp 49-53

[10] C.Chen, T.S.Li, Y.Yeh, C.Chang, "Design and implementation of an adaptive sliding mode dynamic controller for wheeled mobile robots", J. Mechatronics 19 (2009) pp 156-166

[11] Thi Thanh Van Nguyen, Manh Duong Phung, Thuan Hoang Tran, Quang Vinh Tran, "Mobile Robot Localization Using Fuzzy Neural Network Based Extended Kalman Filter", 2012 IEEE International Conference on Control System, Computing and Engineering (ICCSCE), pp.420-425 , Penang, Malaysia, 2012.

[12] T.T. Hoang, P.M. Duong, N.T.T.Van, D.A.Viet, and T.Q. Vinh, Development of an EKF-based localization algorithm using compass sensor and LRF, The 12th International Conference on Control, Automation, Robotics and Vision, ICARCV, Guangzhou - China, 2012.

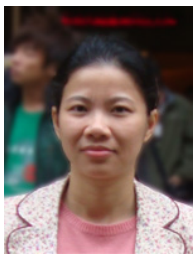

**Nguyễn Thị Thanh Vân** sinh năm 1979. Nhận bằng Cơ điện tử của Viện Công nghệ Châu Á (AIT), Thái Lan năm 2006. Từ năm 2007 đến nay là giảng viên Khoa Điện tử- Viễn thông, Đại học Công nghệ, Đại học Quốc gia Hà Nội. Hướng nghiên cứu chính về các hệ thống điều khiển, điều khiển dẫn đường cho robot di động.

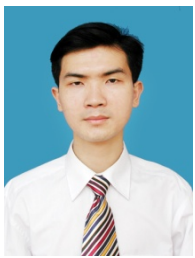

**Phùng Mạnh Dương** nhận bằng cử nhân tại trường Đại học Công nghệ, Đại học Quốc gia Hà Nội năm 2005. Hiện anh là Nghiên cứu sinh tại Khoa Điện tử - Viễn thông, trường Đại học Công Nghệ, Đại Học Quốc Gia Hà Nội. Hướng nghiên cứu chính bao gồm các hệ robot di động phân tán qua mạng máy tính.

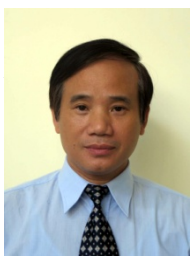

**Trần Quang Vinh** bảo vệ tiến sĩ Vật lý Vô tuyến điện tại ĐH Quốc gia Hà nội trên cơ sở các nghiên cứu thực nghiệm tại Đại học Tổng hợp Kỹ thuật TU Wien (Áo) năm 2001. Hiện là Phó giáo sư, Chủ nhiệm Bộ môn Điện tử và Kỹ thuật máy tính, Trưởng phòng thí nghiệm Các hệ tích hợp thông minh (SIS) tại trường ĐH Công nghệ. Hướng chuyên môn quan tâm hiện nay: Đo lường và điều khiển dùng vi tính và vi xử lý cho các lĩnh vực: vật lý, hóa học, môi trường, y-sinh, nhà thông minh; Điều khiển tự động và robot thông minh (robot di động tự trị, robot nối mạng); Thiết kế chip điện tử tích hợp cỡ lớn VLSI, FPGA, ASIC